\documentclass[lettersize,journal]{IEEEtran}
\usepackage{amsmath,amsfonts}
\usepackage{algorithmic}
\usepackage{algorithm}
\usepackage{array}
\usepackage[caption=false,font=normalsize,labelfont=sf,textfont=sf]{subfig}
\usepackage{textcomp}
\usepackage{stfloats}
\usepackage{url}
\usepackage{verbatim}
\usepackage{graphicx}
\usepackage{cite}
\usepackage{booktabs}
\usepackage{multicol,multirow}
\usepackage{xcolor}

\begin{document}

\title{Improved Hyperspectral Anomaly Detection via Unsupervised Subspace Modeling in the Signed Cumulative Distribution Transform Domain}

\author{Abu Hasnat Mohammad Rubaiyat, Jordan Vincent, Colin Olson
\thanks{A. H. M. Rubaiyat is with Amentum, Hanover, MD 21076, USA (e-mail: abuhasnatmohammad.rubaiyat@us.amentum.com).}
\thanks{J. Vincent is with Tekla Research, Fredericksburg, VA 22407, USA (e-mail:jordan.e.vincent2.ctr@us.navy.mil)}
\thanks{C. Olson is with U.S. Naval Research Laboratory, Washington, DC 20375, USA (e-mail:colin.c.olson.civ@us.navy.mil)}
\thanks{This work was supported by the Office of Naval Research (ONR)} 
}



\maketitle

\begin{abstract}
Hyperspectral anomaly detection (HAD), a crucial approach for many civilian and military applications, seeks to identify pixels with spectral signatures that are anomalous relative to a preponderance of background signatures. Significant effort has been made to improve HAD techniques, but challenges arise due to complex real-world environments and, by definition, limited prior knowledge of potential signatures of interest. This paper introduces a novel HAD method by proposing a transport-based mathematical model to describe the pixels comprising a given hyperspectral image.  In this approach, hyperspectral pixels are viewed as observations of a template pattern undergoing unknown deformations that enables their representation in the signed cumulative distribution transform (SCDT) domain. An unsupervised subspace modeling technique is then used to construct a model of abundant background signals in this domain, whereupon anomalous signals are detected as deviations from the learned model. Comprehensive evaluations across five distinct datasets illustrate the superiority of our approach compared to state-of-the-art methods.
\end{abstract}


\section{Introduction}
\label{sec:intro}
The goal of a hyperspectral anomaly detection (HAD) process is to identify pixels in a hyperspectral image (HSI) exhibiting spectral signatures that are distinct from surrounding background signatures. It has proven to be a powerful tool in many applications including environmental monitoring \cite{xu2019regionally}, intelligent agriculture \cite{goel2003classification}, mineral identification \cite{tan2019parallel}, and aerial searches in defense and surveillance \cite{su2021hyperspectral}, among others. HAD is typically approached as an unsupervised problem, offering the advantage of detecting anomalies without requiring prior information about target or background signatures---a situation that arises frequently in practice. Even when a target spectrum is known, its detection in real-world scenarios can be challenging due to complex and unknown background spectral distributions, varying atmospheres and illuminations, and/or deviations in the assumed emissivity of the target (e.g., due to paint or rust on a surface)---difficulties that can be overcome using a HAD approach. 

Existing HAD methods may be divided into three broad categories: model-based, representation-based, and deep learning-based. A well-known model-based technique is the Reed-Xiaoli (RX) algorithm \cite{reed1990adaptive}, which assumes that the background follows a multivariate Gaussian distribution and detects anomalies by calculating the Mahalanobis distance between target and background pixels. Several improved versions of the RX algorithm include regularized-RX \cite{nasrabadi2008regularization}, kernel-RX \cite{kwon2005kernel}, and segmented-RX \cite{matteoli2010improved}, among others. In real-world applications, however, the assumed statistical distributions underlying these methods are often inaccurate due to the complexity of the background distributions. 

Representation-based methods \cite{liu2012robust,xu2015anomaly} have recently gained popularity as a means to overcome the limitations of model-based techniques. These methods assume that background pixels can be reconstructed from a background dictionary, and anomalies are detected by subtracting the reconstructed background from the original HSI. These approaches are, however, iterative and often require manual parameter tuning. 

Among data-driven deep learning (DL) methods, autoencoders (AEs) and their variants are widely used for HAD \cite{9494034,lu2019exploiting}. Additionally, self-supervised DL techniques have also been explored in recent years \cite{wang2023pdbsnet,wang2023bocknet}. However, the lack of a solid underlying model for the background limits the interpretability and theoretical guarantees of these models, often leading to undesirable outcomes \cite{li2023lrr}.
\begin{table}[tb!]
    \centering
    \normalsize
    \small
    \caption{Description of symbols}
        \begin{tabular}{ll}
        \hline
        Symbols                & Description    \\ \hline
        $s$, $\widehat{s}$ & Spectral signal and its SCDT, respectively \\
        $g$ & Strictly increasing and differentiable function \\
        $\mathcal{T}$ & Set of all possible increasing diffeomorphisms \\
        $\mathbb{S}^{(b)}$/ $\widehat{\mathbb{S}}^{(b)}$ & Set of background signals/SCDTs \\
        $\mathbb{V}^{(b)}$/ $\widehat{\mathbb{V}}^{(b)}$ & Subspace corresponding to $\mathbb{S}^{(b)}$/ $\widehat{\mathbb{S}}^{(b)}$ \\
        $\mathbf{S}^{(b)}$ & Set of background signals present in HSI, $\mathbf{S}$ \\
        $\mathbf{S}^{(a)}$ & Set of anomalous signals present in HSI, $\mathbf{S}$ \\
        \hline
        \end{tabular}
    \label{tab:symbols}
    \vspace{-1em}
\end{table}


In this paper, we introduce an unsupervised technique for hyperspectral anomaly detection that leverages a transport-based modeling approach incorporating the signed cumulative distribution transform (SCDT). Recently, the SCDT—a transport-based transform—has been explored for various estimation and detection tasks \cite{thareja2022signed, Rubaiyat:20, rubaiyat2022nearest}. Originally introduced by Aldroubi et al. \cite{aldroubi2022signed}, the SCDT builds upon the cumulative distribution transform (CDT) \cite{Park:18}, which, based on one-dimensional Wasserstein embedding, was designed to classify strictly positive probability densities. The SCDT generalizes the CDT to handle arbitrary signed signals and possesses several useful properties that enable the application of linear techniques—such as Fisher discriminant analysis, support vector machines, and logistic regression—to nonlinear detection problems in the transform domain \cite{aldroubi2022signed}. A subspace-based classification approach was proposed in \cite{rubaiyat2022nearest}, where each signal class is modeled as a linear subspace, and a test signal is classified based on its nearest subspace. This concept was further expanded in \cite{rubaiyat2024end}, where signal classes are modeled as unions of subspaces, and classification is performed in the SCDT domain by finding the nearest local subspace to the test signal.

While prior subspace search methods have been developed primarily for supervised classification, this paper presents a novel unsupervised subspace modeling framework specifically for hyperspectral anomaly detection (HAD). We propose a transport-based mathematical model for the hyperspectral pixels, where the spectral signals corresponding to the pixels are considered as observations of a template pattern undergoing certain unknown deformations such as translation, dispersion, warping, and others.
These deformations, often caused by factors such as shadowing, look angles, atmospheric absorption, and similar effects, are commonly seen in spectral data which makes the proposed model well suited for hyperspectral signal analysis.
Moreover, this model induces convexity of the hyperspectral pixel representation in the SCDT space. This property facilitates the construction of a subspace model that characterizes the dominant background signals. Anomalies are then detected as signals that deviate from the learned subspace in the transform domain.

Extensive experimental results highlight the superior performance of our approach compared to traditional and state-of-the-art machine learning-based methods. In addition, the proposed technique is non-iterative and does not require manual parameter tuning which makes it more suitable for practical, real-world applications. The remaining of this paper is structured as follows. Section \ref{sec:preli} introduces the mathematical notation used throughout the manuscript and provides a brief review of the definition and key properties of the SCDT. In Section \ref{sec:method}, we formulate the unsupervised HAD problem based on the transport-based model and present the proposed solution. Section \ref{sec:exp_res} describes the datasets and presents the experimental results, along with discussions of the findings. Finally, Section \ref{sec:conc} presents the conclusions of the study.

\section{Preliminaries}
\label{sec:preli}

\subsection{Notation}
\label{sec:notation}

Throughout the manuscript, we work with 1-D spectral signals $s$ such that $\int_{\Omega_s}|s(u)|du<\infty$, where $\Omega_s\subseteq\mathbb{R}$ is the domain over which $s$ is defined. $\|.\|_{L_1}$ and $\|.\|_{L_2}$ denote the $L_1$ and $L_2$ norms, respectively. Some symbols used throughout the manuscript are listed in Table \ref{tab:symbols}.

\subsection{Signed Cumulative Distribution Transform}
\label{sec:scdt}
The signed cumulative distribution transform (SCDT), introduced in \cite{aldroubi2022signed}, extends the cumulative distribution transform (CDT) \cite{Park:18}. The CDT is a nonlinear 1D signal transform from the space of smooth positive probability densities to the space of diffeomorphisms. Consider a signal $s(x)$ and a reference signal $s_0(y)$, where both signals are positive within their respective domains $\Omega_s$ and $\Omega_{s_0}$, and $\int_{\Omega_s}s(u)du = \int_{\Omega_{s_0}}s_0(u)du = 1$.
The CDT of $s(x)$, denoted as $s^*(y)$, is the function that satisfies the following equation:
\begin{equation}
    \int_{\inf(\Omega_s)}^{s^*(y)} s(u)du = \int_{\inf(\Omega_{s_0})}^{y} s_0(u)du.
    \label{eq:cdt}
\end{equation}
Alternatively, $s^*(y)$ can be expressed as $s^*(y) = S^{-1}(S_0(y))$,
where $S(x) = \int_{-\infty}^{x} s(u)du$ and $S_0(y) = \int_{-\infty}^{y} s_0(u)du$.
For a uniform reference signal, this simplifies to:
\begin{equation}
	s^*(y) = S^{-1}(y).
	\label{eq:cdt_alt}
\end{equation}

The SCDT extends the CDT to general finite signed signals. For a non-negative signal $s(x)$ with arbitrary mass, the SCDT is defined as:
\begin{equation}
    \widehat{s}(y) = \begin{cases}
    \left(s^*(y),\|s\|_{L_1}\right),& \text{if } s\neq 0\\
    (0,0),              & \text{if } s=0,
\end{cases}
\label{eq:scdt_mass}
\end{equation}
where $s^*$ is the CDT (Eq.~(\ref{eq:cdt_alt})) of the normalized signal $\frac{s}{\|s\|_{L_1}}$. For a signed signal $s(x)$, the signal is first decomposed as $s(x) = s^+(x) - s^-(x)$, where $s^+(x)$ and $s^-(x)$ are the absolute values of the positive and negative parts of the signal $s(x)$. The SCDT of $s(x)$ is then given by:
\begin{equation}
    \widehat{s}(y) = \left(\widehat{s}^+(y), \widehat{s}^-(y)\right),
    \label{eq:scdt}
\end{equation}
where $\widehat{s}^+(y)$ and $\widehat{s}^-(y)$ are the transforms, as defined in Eq.~(\ref{eq:scdt_mass}), of $s^+(x)$ and $s^-(x)$, respectively. The SCDT has several properties that can simplify various signal processing problems, as demonstrated in recent studies \cite{rubaiyat2024end}.

\textbf{Composition property:}
The SCDT of the signal $s_g=g's\circ g$ is given by \cite{aldroubi2022signed}:
\begin{equation}
    \widehat{s}_g = \left(g^{-1}\circ (s^+)^*,\|s^{+}\|_{L_1},g^{-1}\circ (s^-)^*,\|s^{-}\|_{L_1}\right),
    \label{eq:scdt_composition}
\end{equation}
where, $s\circ g = s(g(x))$ is the composition of $s(x)$ with $g(x)$, $g(x)$ is an invertible, differentiable, monotonically increasing function, and $g'=dg(x)/dx$. 
For example, in case of translation and scaling: $g(x) = \omega x - \mu$, and the signal $s_g(x)=\omega s(\omega x - \mu)$. The SCDT of $s_g$ is then derived as: 
\begin{equation*}
    \widehat{s}_g = \left(\frac{(s^+)^* + \mu}{\omega},\|s^{+}\|_{L_1},\frac{(s^-)^* + \mu}{\omega},\|s^{-}\|_{L_1}\right).
\end{equation*}
This composition property indicates that changes in the independent variable due to $g(x)$ affect only the dependent variable in the SCDT domain, making it useful for modeling certain nonlinear deformations in the original signal space linearly in the SCDT space.

\textbf{Convexity property:}
Given a fixed signal $\varphi$, and a set $\mathcal{G}\subset\mathcal{T}$ of 1D deformations (such as warping, scaling, etc.), the $(\varphi, \mathcal{G})$-associated set of signals $\mathbb{S_{\varphi,\mathcal{G}}}$ is defined as: $\mathbb{S_{\varphi, \mathcal{G}}} =\{s_j|s_j=g'_j\varphi\circ g_j, g_j\in \mathcal{G}\}$. According to the convexity property, the set $\widehat{\mathbb{S}}_{\varphi, \mathcal{G}} = \{\widehat{s}_j:s_j\in \mathbb{S}_{\varphi, \mathcal{G}}\}$ is convex for any $\varphi$ if and only if the set $\mathcal{G}^{-1}=\{g_j^{-1}:g_j\in \mathcal{G}\}$ is convex \cite{aldroubi2022signed}.

The set $\mathbb{S}_{\varphi, \mathcal{G}}$ can be viewed as an algebraic model for a collection of signal data, such as background data points in a HSI, with $\varphi$ serving as a prototype signal or template. For simplicity, in the remaining text we suppress the ${\varphi, \mathcal{G}}$ subscript in the notation for $\mathbb{S}$ wherever appropriate. Next, we demonstrate how the properties of the SCDT simplify the hyperspectral anomaly detection problem.

\section{Proposed Method}
\label{sec:method}

\begin{figure}[!tb]
    \centering
    \includegraphics[width=0.48\textwidth]{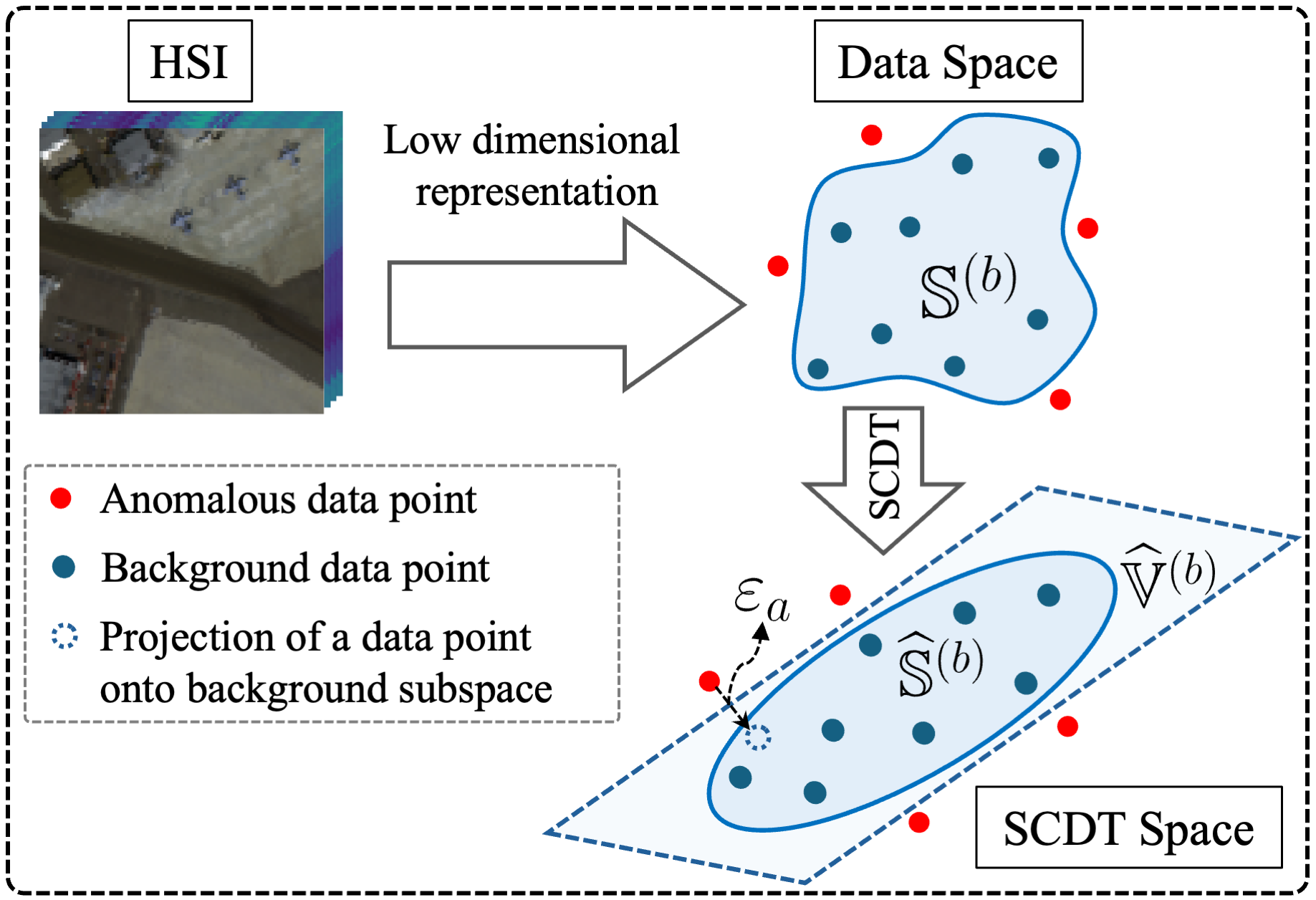}
    \caption{ The proposed approach begins by estimating the background subspace $\widehat{\mathbb{V}}^{(b)}$ in the SCDT domain, which spans the set $\widehat{\mathbb{S}}^{(b)}$. The anomaly score $\varepsilon_a$ for a test data point is then computed based on its subspace distance from $\widehat{\mathbb{V}}^{(b)}$.}
    \label{fig:algo}
    \vspace{-1em}
\end{figure}

\subsection{Transport-based Model and Problem Statement}
\label{sec:problem}
This section introduces a transport-based model for hyperspectral image (HSI) pixels and formulates the hyperspectral anomaly detection (HAD) problem. Consider a HSI, $\mathbf{S}=\{s_i\}_{i=1}^N\in\mathbb{R}^{D\times N}$, where $N$ denotes the number of pixels and $D$ represents the number of spectral bands, with each $s_i\in\mathbb{R}^D$. The pixels in $\mathbf{S}$ can be divided into two categories: background (non-anomalous) and anomalous. In HSI, there usually exists strong correlation among the background pixels. In this study, we model the spectral signals corresponding to the background pixels as the instances of a certain template pattern observed under some unknown deformations, such as dispersion, warping, or similar transformations. The background signals are thus described using the following mathematical model:

\textbf{Background signal Model: }
Let  $\mathcal{G}\subset\mathcal{T}$ denotes a set of increasing 1-D deformations of a special kind (e.g., translation, dispersion, warping, etc.). The 1D transport-based model for the set of background pixels is defined as:
\begin{align}
    &\mathbb{S}^{(b)} = \left\{s_j^{(b)}|s_j^{(b)}=g'_{j}\varphi^{(b)}\circ g_j, g_j\in\mathcal{G}\right\}, \nonumber \\
    &\mathcal{G}^{-1}= \left\{\sum_{i=1}^k\alpha_i f_i,\,\,\, \alpha_i\geq 0\right\},
    \label{eq:bg_model}
\end{align}
where $s^{(b)}_j$ is a background data point, $\varphi^{(b)}$ is the template spectrum for the background, $\{f_1, ..., f_k\}$ denotes a set of linearly independent and strictly increasing functions, and $k$ is a positive integer.

\textbf{Anomaly Detection Problem: }
Given a hyperspectral image $\mathbf{S}$ with a large number of background pixels $\mathbf{S}^{(b)}\subseteq\mathbb{S}^{(b)}$ and a small number of anomalous pixels $\mathbf{S}^{(a)}$, where $\mathbf{S}^{(a)}\cap\mathbb{S}^{(b)}=\emptyset$ and $\mathbb{S}^{(b)}$ is defined by the model in Eq.~(\ref{eq:bg_model}), the objective is to determine whether a data point $s$, drawn from $\mathbf{S}$, is anomalous, that is $s\notin\mathbb{S}^{(b)}$.

\subsection{Proposed Solution}
The mathematical model for background pixels, as defined in Eq.~(\ref{eq:bg_model}), generally leads to a nonlinear and non-convex signal class, which makes it difficult for traditional subspace learning methods to accurately represent the background. Applying the SCDT can simplify the data geometry \cite{li2022geodesic}, making it more amenable to modeling nonlinear physical systems \cite{rubaiyat2024data}. In addition, the SCDT’s composition and convexity properties facilitate solving nonlinear estimation and detection problems more easily in the SCDT domain compared to the original signal space \cite{rubaiyat2024end,thareja2022signed}. Therefore, in this work, we first transform the spectral signals to the SCDT space and then address the anomaly detection problem in the transform domain.
By leveraging the composition property of the SCDT, the background model in the SCDT domain is given by:
\begin{align}
    &\widehat{\mathbb{S}}^{(b)} = \left\{\widehat{s}_j^{(b)}| \widehat{s}_j^{(b)}= g_j^{-1}\circ \widehat{\varphi}^{(b)},
    g_j\in \mathcal{G}\right\}, \nonumber \\
    &\mathcal{G}^{-1}= \left\{\sum_{i=1}^k\alpha_i f_i,\,\,\, \alpha_i\geq 0\right\},
    \label{eq:bg_model_scdt}
\end{align}
where $g_j^{-1}\circ \widehat{\varphi}^{(b)}$ is the SCDT of the signal $g'_{j}\varphi^{(b)}\circ g_j$. Here, we define $\mathcal{G}^{-1}$ in such a way that it becomes convex by definition. In many practical scenarios, signal sets of interest can be generated through such differentiable, strictly increasing deformations, for example, $g^{-1}(x) = \sum_ip_ix^i,\,g'>0$. Therefore, the set $\widehat{\mathbb{S}}^{(b)}$ is convex according to the convexity property of the SCDT.

To formulate the solution to the anomaly detection problem defined above, we adopt a subspace-based approach. The subspace associated with the background pixels can be described in the SCDT domain as follows:
\begin{equation}
    \widehat{\mathbb{V}}^{(b)} = \text{Span}\left( \widehat{\mathbb{S}}^{(b)} \right).
    \label{eq:bg_subspace}
\end{equation}
If a test signal $s\in\mathbf{S}$ corresponds to the background, its SCDT $\widehat{s}$ should lie within the subspace $\widehat{\mathbb{V}}^{(b)}$, resulting in a subspace distance ideally equal to zero. Conversely, for an anomalous signal, this distance in the SCDT domain will be strictly greater than zero. Following this principle, the anomaly score for a test data point $s$ drawn from $\mathbf{S}$ is computed as:
\begin{equation}
    \varepsilon_a = d^2(\widehat{s},\widehat{\mathbb{V}}^{(b)}),
    \label{eq:score_anom}
\end{equation}
where, $d(.,.)$ represents the Euclidean distance between $\widehat{s}$ and the nearest point in $\widehat{\mathbb{V}}^{(b)}$. Since $\widehat{\mathbb{V}}^{(b)}$ characterizes the background subspace, $\varepsilon_a$ will be higher for anomalous data points compared to the background pixels. It is important to note that although the proposed model assumes the background signals are observations of a template pattern $\varphi^{(b)}$ under confounding deformations $\mathcal{G}$, the knowledge of either $\varphi^{(b)}$ or $\mathcal{G}$ is not necessary for the solution.


\subsection{Algorithm: Anomaly Detection in SCDT Domain}
\label{sec:alg}
The proposed algorithm begins by applying the SCDT transform to the spectral signals of the HSI denoted as $\mathbf{S}$, resulting in $\widehat{\mathbf{S}}$. Since this is an unsupervised method, the signals corresponding to the background pixels are unknown, and therefore, it is not possible to directly estimate $\widehat{\mathbb{V}}^{(b)}$ using Eq.~(\ref{eq:bg_subspace}). Let us assume that the background signals in $\mathbf{S}$ are generated according to the model $\mathbb{S}^{(b)}$. According to Eq.~(\ref{eq:bg_model_scdt}), $\mathbb{S}^{(b)}$ can be defined in the SCDT domain as:
\begin{equation}
    \widehat{\mathbb{S}}^{(b)} = \left\{\left(\sum_{i=1}^k\alpha_if_i \right)\circ \widehat{\varphi}^{(b)},\,\alpha_i\geq 0 \right\},
\end{equation}
where $\{f_1, ..., f_k\}$ denotes a set of $k$ linearly independent increasing functions. This definition indicates that the background subspace $\widehat{\mathbb{V}}^{(b)}$ is a $k$-dimensional space. To estimate this subspace, we apply principal component analysis (PCA) to $\widehat{\mathbf{S}}$ and select the $k$ most significant principal components to obtain the basis vectors that span the space $\widehat{\mathbb{V}}^{(b)}$. In practical scenarios, the parameter $k$ is typically unknown, and it must be chosen such that the estimated subspace captures all potential confounding sources of variability present in the background signals. In this work, we select $k$ such that the cumulative variance explained by the first $k$ principal components accounts for at least $99.99\%$ of the total variance in the data.. We then construct the matrix $B$ with the computed basis vectors in its column: $B=\left[b_1, ..., b_k \right]$. The anomaly score for a given test signal $s$ is calculated as:
\begin{equation}
    \varepsilon_a=\|\widehat{s} - BB^T\widehat{s}\|^2_{L_2},
    \label{eq:score_anom_alg}
\end{equation}
where $BB^T$ is the orthogonal projection matrix onto the subspace spanned by the columns of $B$. 

Note that the matrix $B$ contains the basis vectors that span the background-dominated subspace of the HSI $\mathbf{S}$. Thus, the projection $BB^T\widehat{s}$ can be interpreted as the reconstruction of the SCDT of the test signal $s$ using only the components within the background subspace. Consequently, the anomaly score $\varepsilon_a$, as defined in Eq.~(\ref{eq:score_anom_alg}), represents the reconstruction error between the test signal $\widehat{s}$ and its projection onto the estimated background subspace. Therefore, an anomalous signal $s$ is expected to yield higher values of $\varepsilon_a$ compared to background pixels. The overall algorithm is illustrated in Fig.~\ref{fig:algo}.


\begin{figure}[!tb]
    \centering
    \includegraphics[width=0.48\textwidth]{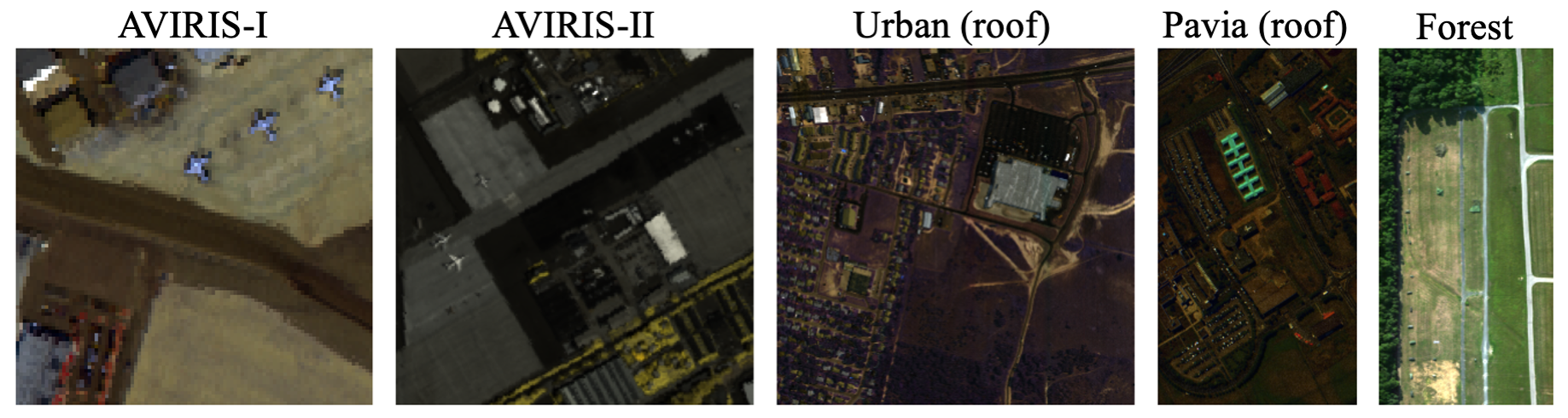}
    \caption{Approximate RGB images of the datasets.}
    \label{fig:dataset}
    \vspace{-1em}
\end{figure}

\begin{figure*}[!tb]
    \centering
    \includegraphics[width=0.48\textwidth]{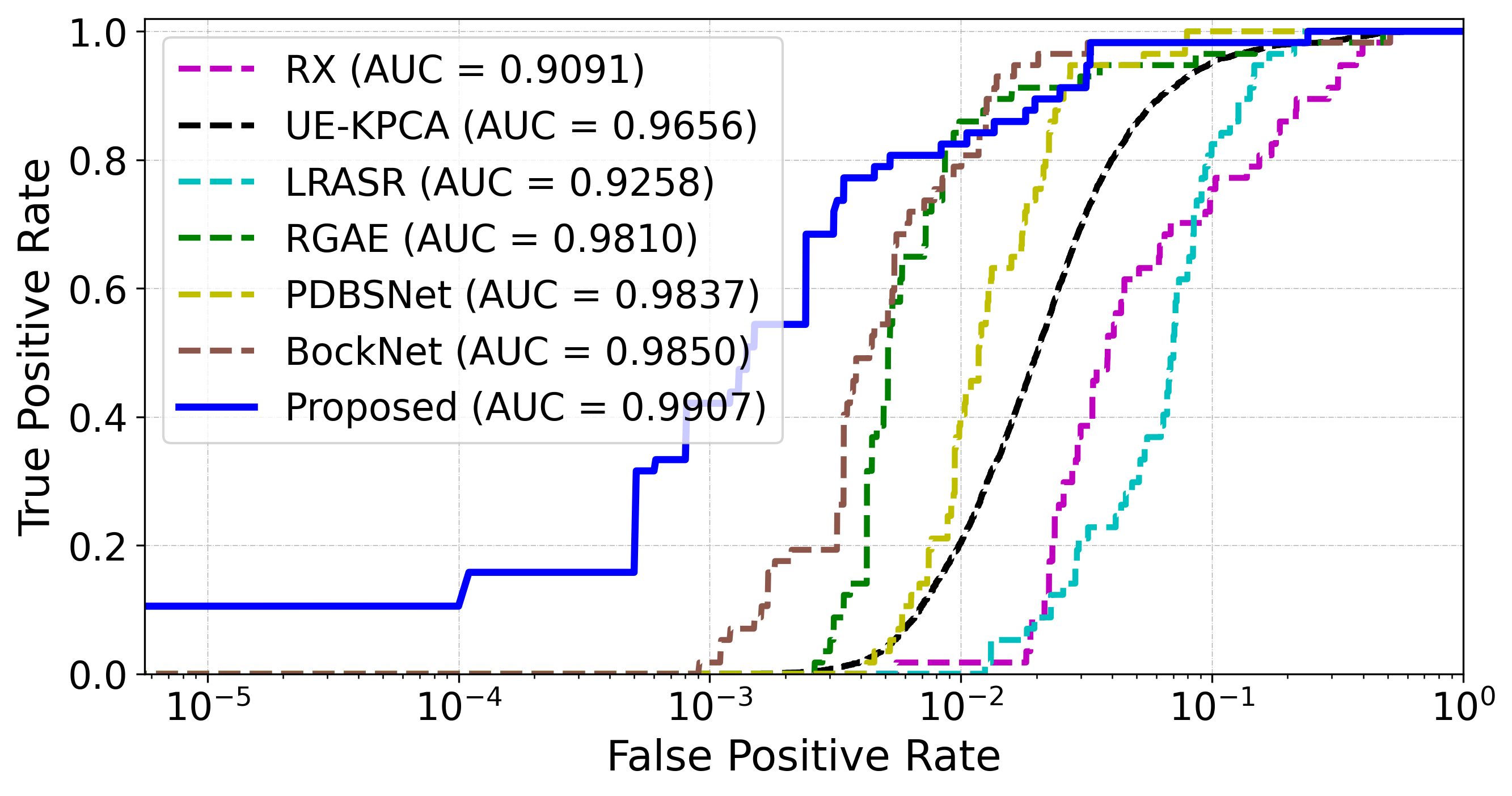}
    \includegraphics[width=0.48\textwidth]{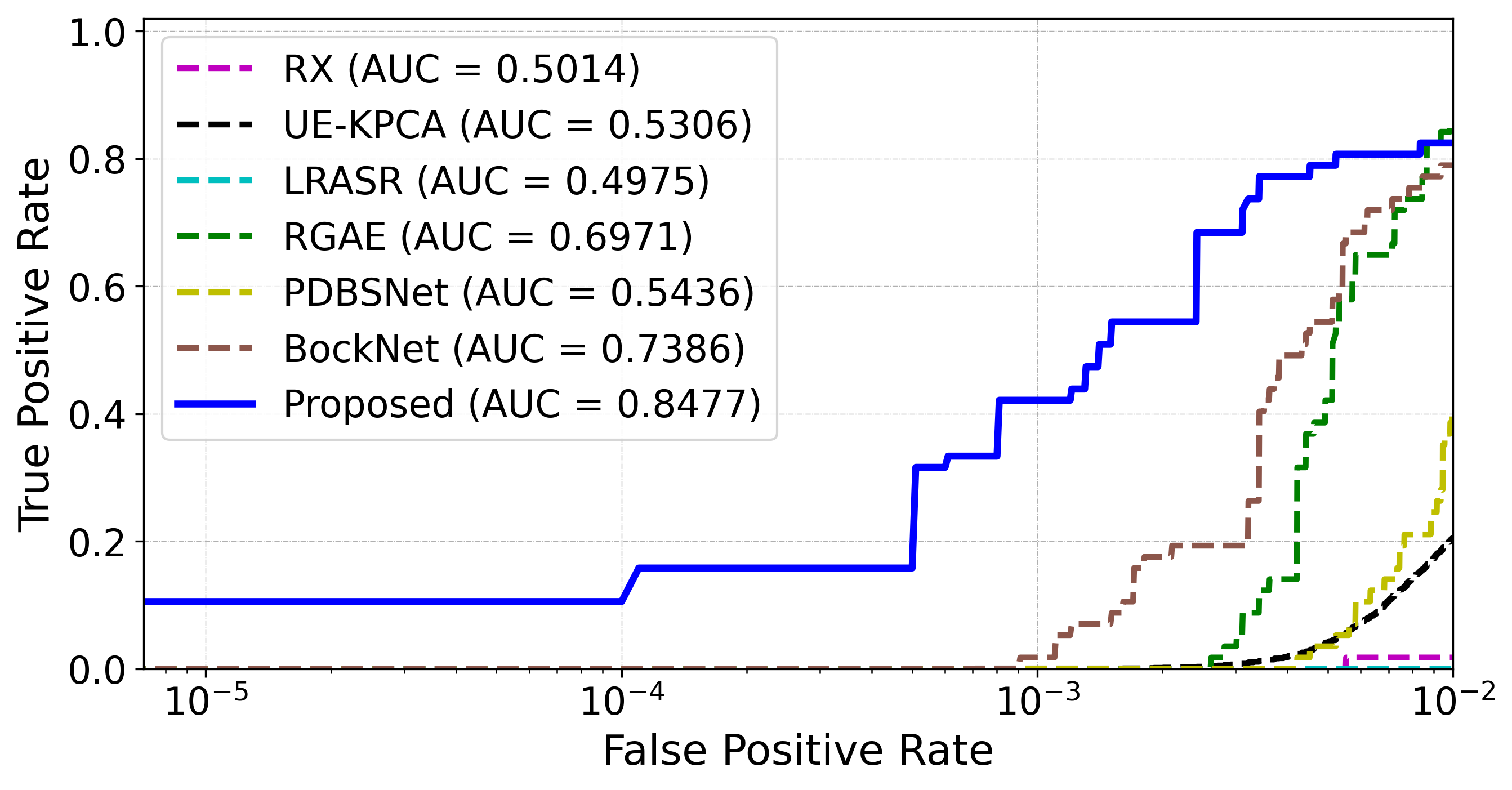}
    \vspace{-1em}
    \caption{Experimental results on AVIRIS-I dataset: (left) full ROC curves up to $\text{FPR}= 10^{0}$, and (right) partial ROC curves restricted to $\text{FPR}\leq 10^{-2}$.}
    \label{fig:roc_avirisI}
\end{figure*}

\begin{figure*}[!tb]
    \centering
    \includegraphics[width=0.48\textwidth]{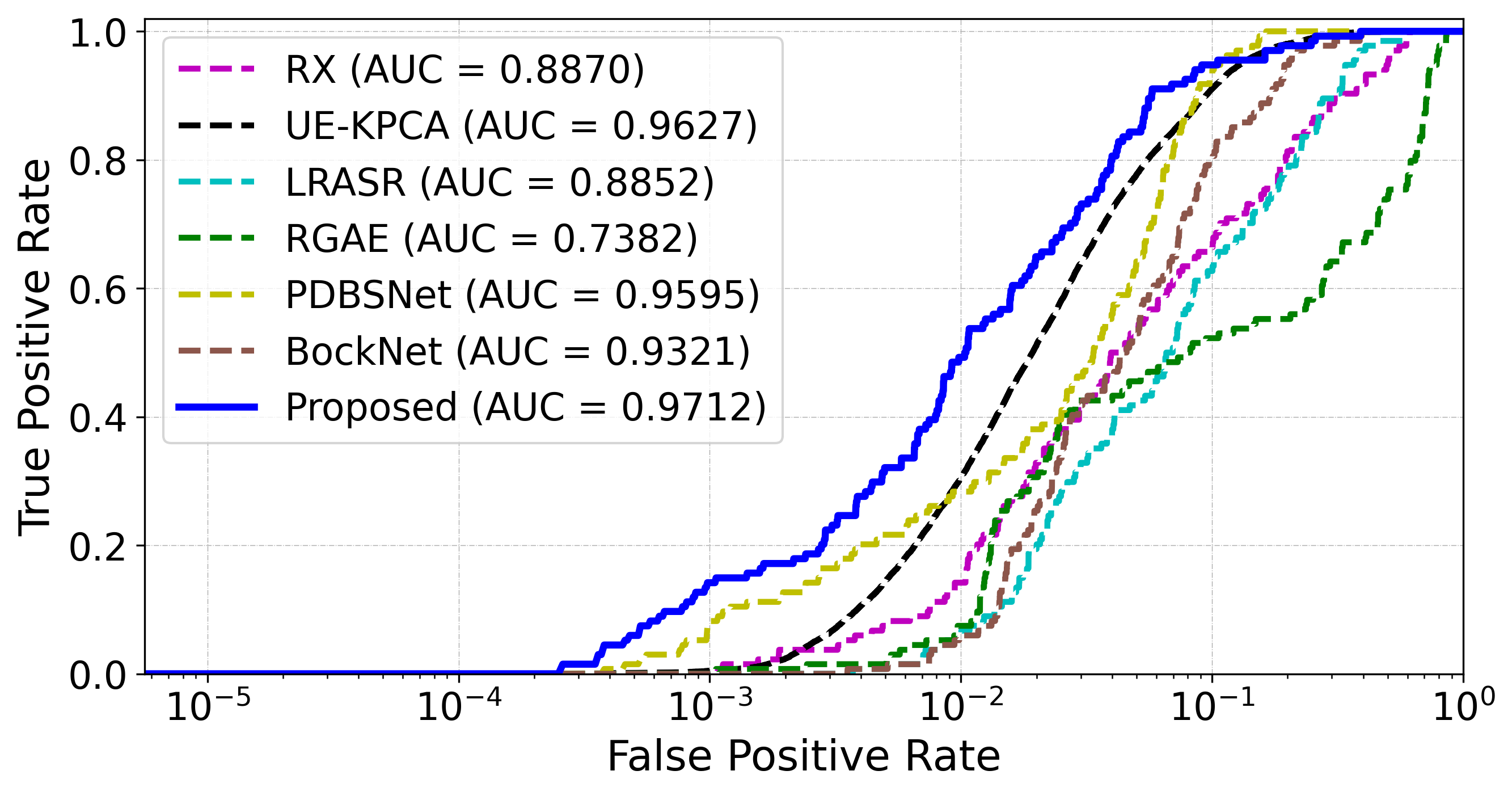}
    \includegraphics[width=0.48\textwidth]{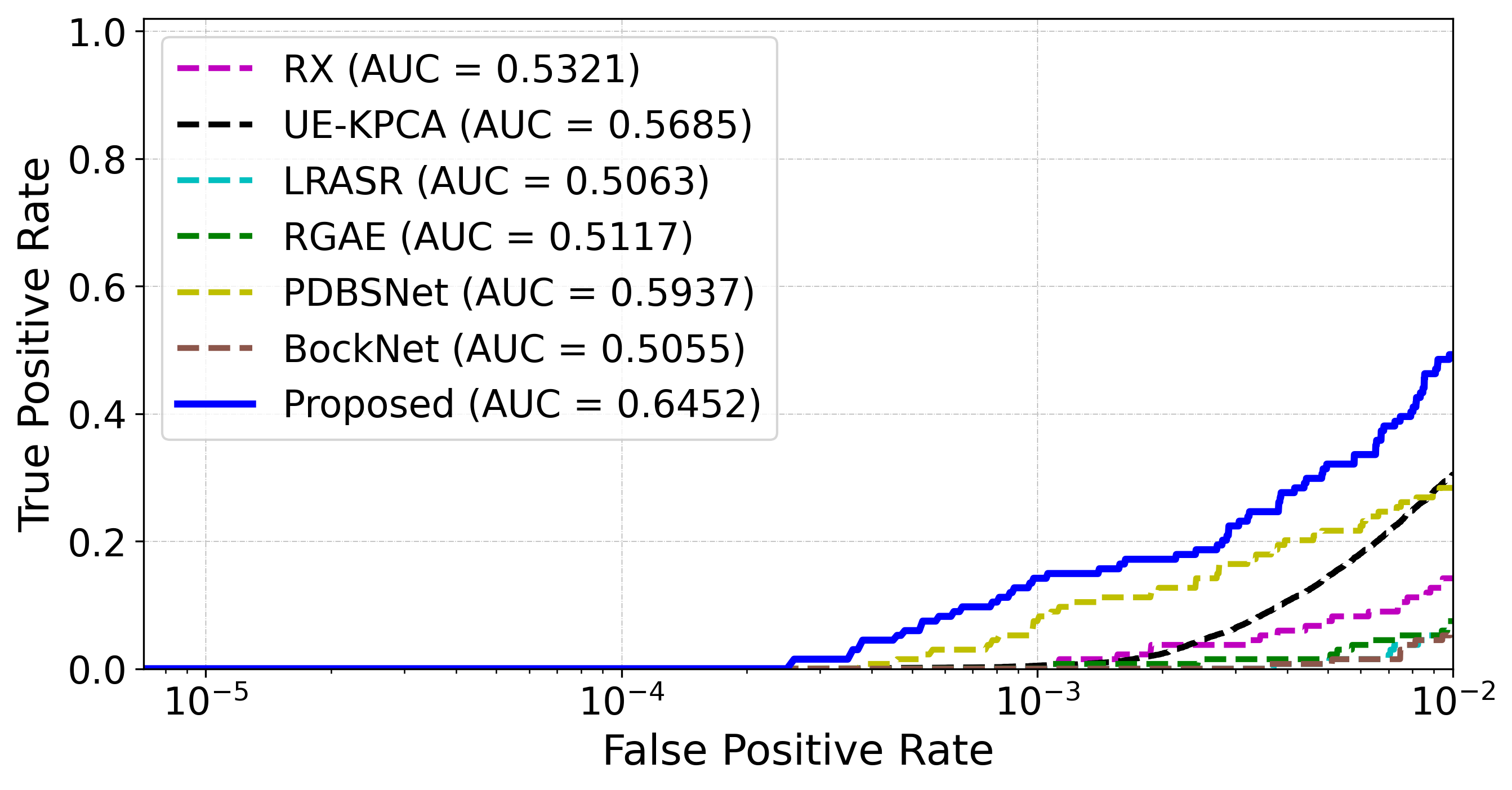}
    \vspace{-1em}
    \caption{Experimental results on AVIRIS-II dataset: (left) full ROC curves up to $\text{FPR}= 10^{0}$, and (right) partial ROC curves restricted to $\text{FPR}\leq 10^{-2}$.}
    \label{fig:roc_avirisII}
\end{figure*}

\begin{figure*}[!tb]
    \centering
    \includegraphics[width=0.48\textwidth]{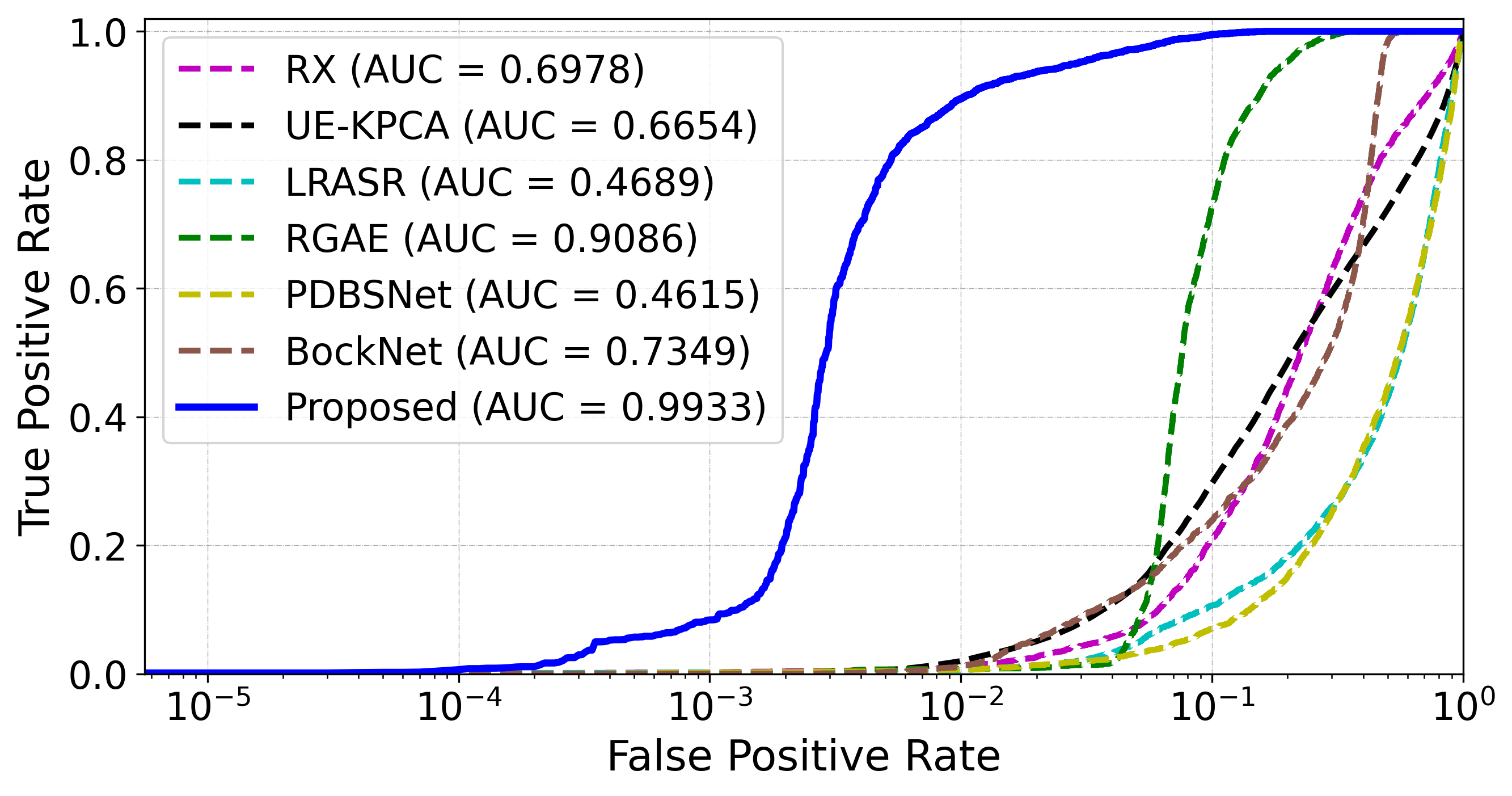}
    \includegraphics[width=0.48\textwidth]{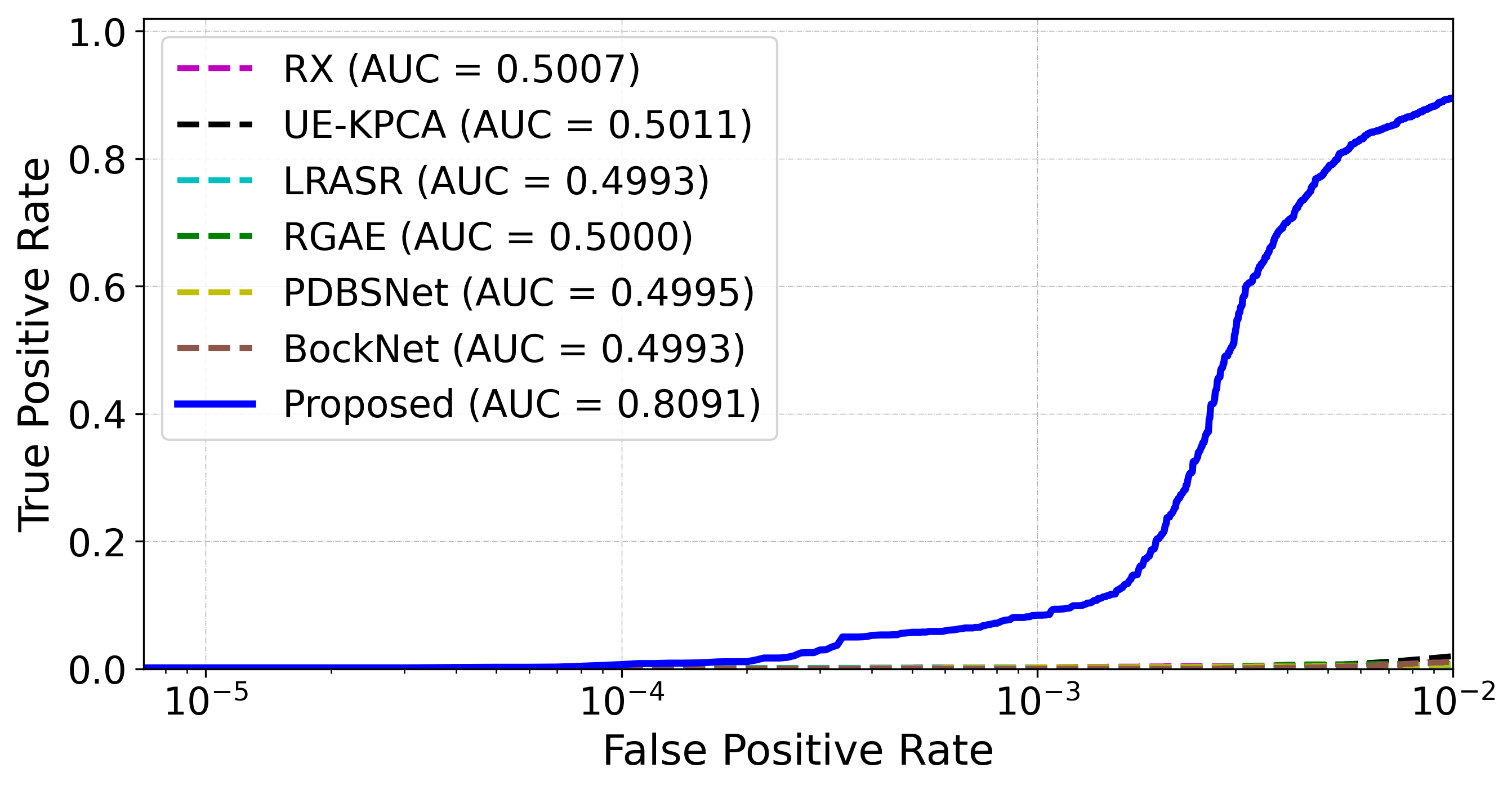}
    \vspace{-1em}
    \caption{Experimental results on Urban dataset: (left) full ROC curves up to $\text{FPR}= 10^{0}$, and (right) partial ROC curves restricted to $\text{FPR}\leq 10^{-2}$.}
    \label{fig:roc_urban}
\end{figure*}

\begin{figure*}[!tb]
    \centering
    \includegraphics[width=0.48\textwidth]{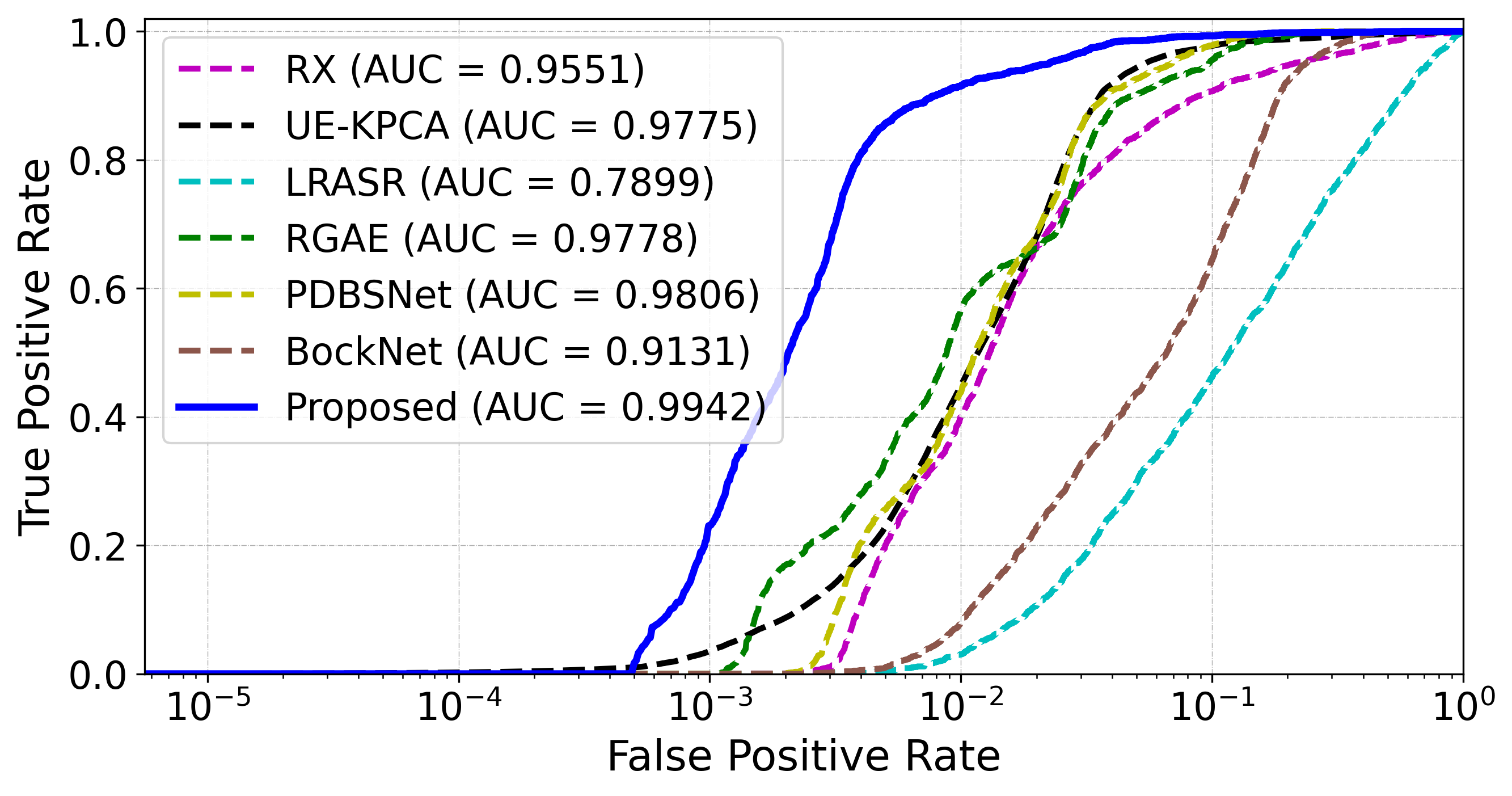}
    \includegraphics[width=0.48\textwidth]{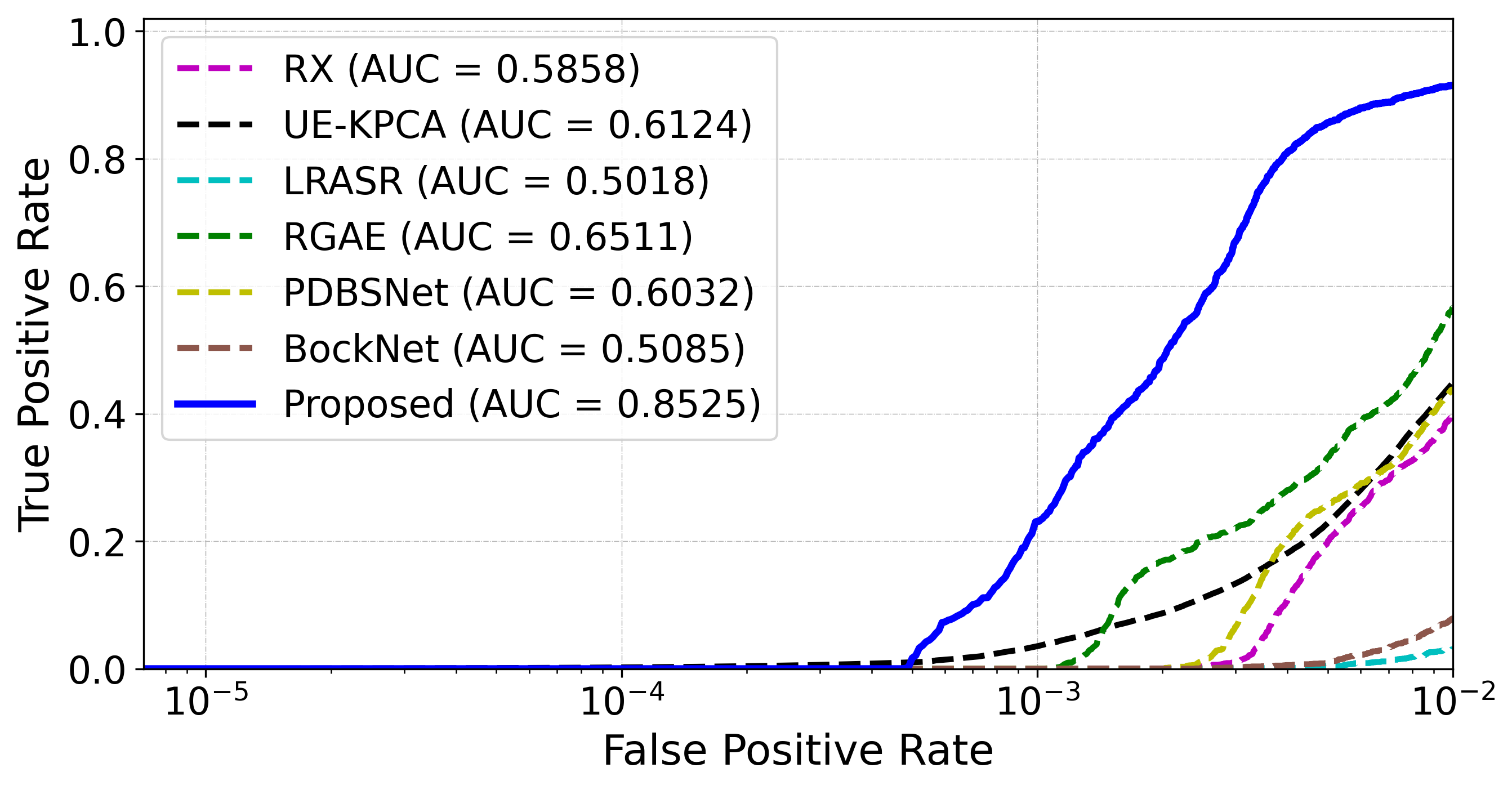}
    \vspace{-1em}
    \caption{Experimental results on Pavia dataset: (left) full ROC curves up to $\text{FPR}= 10^{0}$, and (right) partial ROC curves restricted to $\text{FPR}\leq 10^{-2}$.}
    \label{fig:roc_pavia}
\end{figure*}

\begin{figure*}[!tb]
    \centering
    \includegraphics[width=0.48\textwidth]{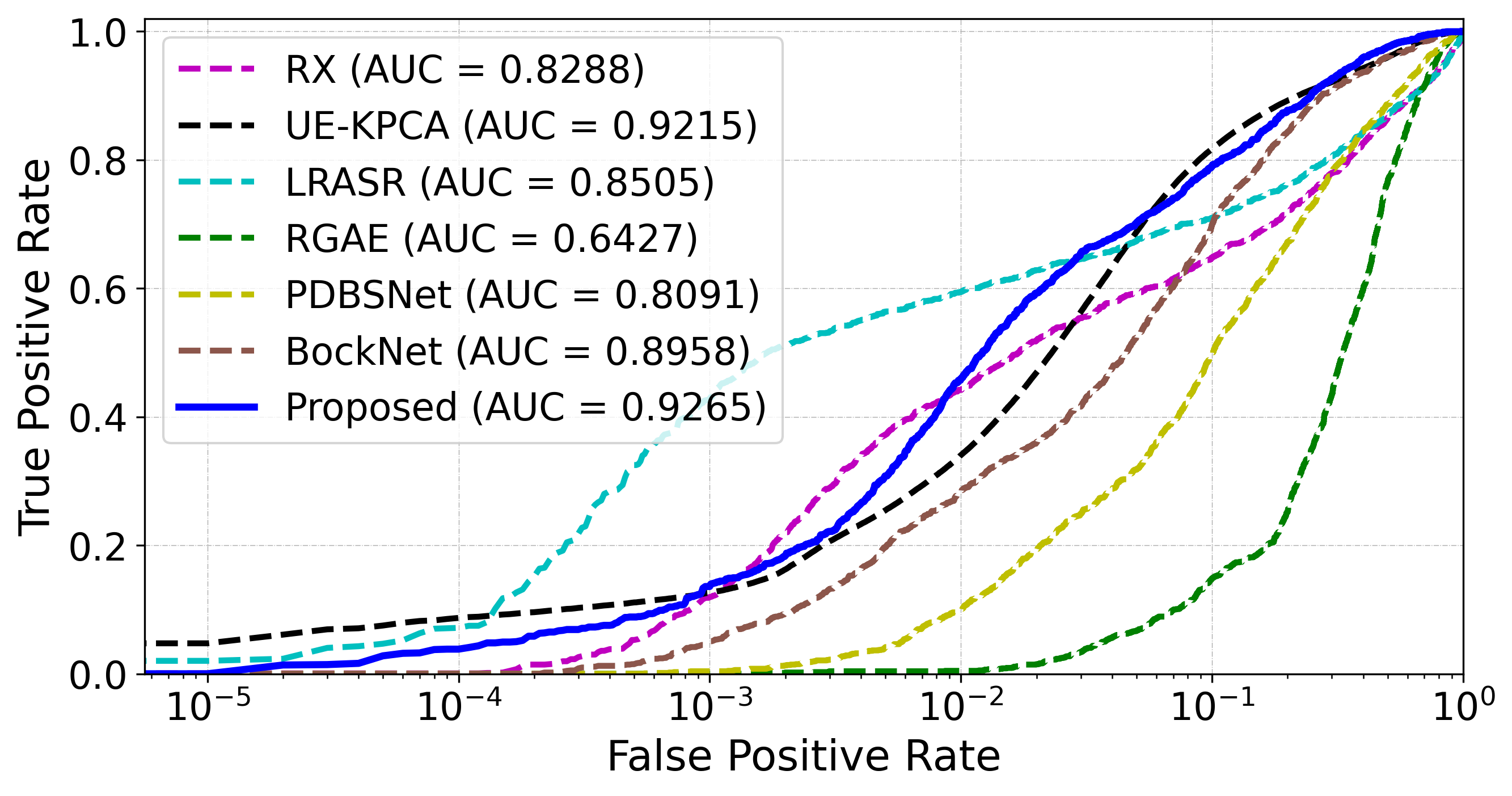}
    \includegraphics[width=0.48\textwidth]{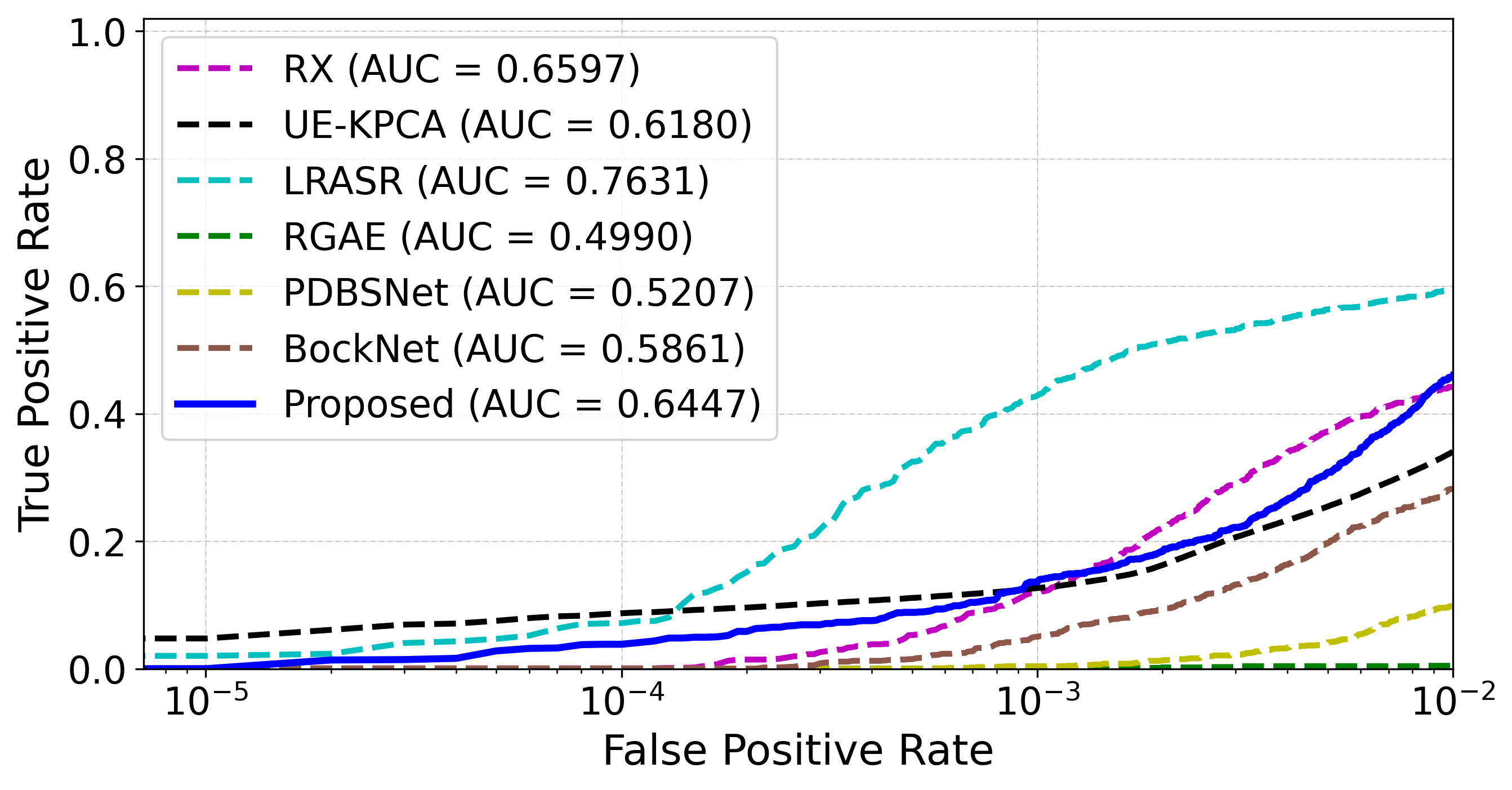}
    \vspace{-1em}
    \caption{Experimental results on Forest dataset: (left) full ROC curves up to $\text{FPR}= 10^{0}$, and (right) partial ROC curves restricted to $\text{FPR}\leq 10^{-2}$.}
    \label{fig:roc_forest}
\end{figure*}

\begin{table*}[tb!]
\centering
\renewcommand{\arraystretch}{1.2}
\setlength{\tabcolsep}{4.0pt}
\caption{Comparative AUC scores integrated up to FPR values equal to $10^{-3}$, $10^{-2}$, and $10^{0}$ for five datasets.}
\begin{tabular}{|l|ccc|ccc|ccc|ccc|ccc|}
\hline
\textbf{Methods}
& \multicolumn{3}{c|}{\textbf{AVIRIS-I}} 
& \multicolumn{3}{c|}{\textbf{AVIRIS-II}} 
& \multicolumn{3}{c|}{\textbf{Urban}} 
& \multicolumn{3}{c|}{\textbf{Pavia}} 
& \multicolumn{3}{c|}{\textbf{Forest}} \\ \cline{1-16}
\textbf{FPR $\leq$}
& $10^{-3}$ & $10^{-2}$ & $10^{0}$ 
& $10^{-3}$ & $10^{-2}$ & $10^{0}$ 
& $10^{-3}$ & $10^{-2}$ & $10^{0}$ 
& $10^{-3}$ & $10^{-2}$ & $10^{0}$ 
& $10^{-3}$ & $10^{-2}$ & $10^{0}$ \\ \hline\hline

RX       & 0.4997 &  0.5014  & 0.9091   & 0.4997 &  0.5321 &  0.8870 & 0.5004 &  0.5007 & 0.6978  & 0.4997 & 0.5858 & 0.9551       & 0.5270 & 0.6597 & 0.8288 \\
UE-KPCA  & 0.4998 &  0.5306  & 0.9656  & 0.5005 & 0.5685  &  0.9627  & 0.4999 & 0.5011 & 0.6654  & 0.5064 & 0.6124 &  0.9775      & 0.5534 & 0.6180 & 0.9215 \\
LRASR    & 0.4997 &  0.4975  &  0.9258  & 0.4997 &  0.5063 &  0.8852  & 0.5005 &  0.4993  & 0.4689 & 0.4997 & 0.5018 &   0.7899     & \textbf{0.6411} & \textbf{0.7631} & 0.8505 \\
RGAE     & 0.4997 &  0.6971  & 0.9810  & 0.4997 &  0.5117 &  0.7382 & 0.4998 & 0.5000 &  0.9086 &   0.4997     &  0.6511      &  0.9778      & 0.4997 & 0.4990 & 0.6427 \\
PDBSNet  & 0.4997 &  0.5436  &  0.9837  & 0.5103 & 0.5937 &  0.9595 & 0.5002 &  0.4995  &  0.4615 &    0.4997    &   0.6032   &   0.9806   & 0.5003 & 0.5207 & 0.8091 \\
BockNet  & 0.5006 &  0.7386  &  0.9850  & 0.4997 & 0.5055 &  0.9321 & 0.4997 &  0.4993 &  0.7349 &  0.4997    &  0.5085    &  0.9131    & 0.5094 & 0.5861 & 0.8958 \\
Proposed & \textbf{0.6274} &  \textbf{0.8477} & \textbf{0.9907}  & \textbf{0.5295} &  \textbf{0.6452} &  \textbf{0.9712} & \textbf{0.5235} &   \textbf{0.8091} &  \textbf{0.9933} & \textbf{0.5297}  & \textbf{0.8525}  & \textbf{0.9942}  & 0.5419 & 0.6447 & \textbf{0.9265} \\

\hline
\end{tabular}
\label{tab:auc_scores}
\vspace{-1.0em}
\end{table*}

\section{Experiments and Results}
\label{sec:exp_res}
\subsection{Datasets}
\label{sec:data}
Five datasets are used to evaluate our method. The first dataset, AVIRIS-I \cite{xu2015anomaly}, was captured by the Airborne Visible/Infrared Imaging Spectrometer (AVIRIS) over San Diego, CA, USA. It consists of 224 spectral bands, with 186 bands retained for the experiments after removing some bands due to low SNR. The full AVIRIS image measures $400\times 400$ pixels, but following \cite{xu2015anomaly}, a $100\times 100$ region from the upper-left corner of the scene is selected for analysis. 
The second dataset, AVIRIS-II \cite{huyan2018hyperspectral}, is a $200\times 200$ area selected from the AVIRIS image. In contrast to AVIRIS-I, this dataset includes a wider variety of background materials. 
In both datasets, the pixels corresponding to the airplanes are considered anomalous. 

The third dataset \cite{xu2015anomaly} is a Hyperspectral Digital Imaging Collection Experiment (HYDICE) dataset acquired from an airborne platform. It covers an urban area containing vegetation, construction sites, and roads with some vehicles. In this dataset, the pixels corresponding to roofs are designated as anomalous targets. The dataset has a total size of $307 \times 307$ pixels, with 162 bands retained after removing low-SNR and water vapor absorption bands. 

The fourth dataset used is the Pavia University HSI \cite{grana2021hyperspectral}, acquired with the ROSIS (Reflective Optics System Imaging Spectrometer) sensor. It consists of 103 spectral bands with a spatial resolution of $610 \times 340$ pixels, capturing a hyperspectral aerial view of the Pavia University area. The scene includes various materials such as asphalt, meadows, trees, and metal sheets. In this study, pixels corresponding to the painted metal sheets are considered as anomalous targets.

The final dataset is the HYDICE Forest Radiance I \cite{olson2017novel}, which contains 210 spectral bands. After excluding bands with low signal-to-noise ratio (SNR), 158 bands are retained for analysis. While the original image measures $1280 \times 308$ pixels, a $600 \times 293$ subregion is selected for evaluation. Approximate RGB images of all datasets are presented in Fig.~\ref{fig:dataset}.

\subsection{Results and Discussion}
\label{sec:res}
We conducted anomaly detection experiments on five hyperspectral datasets and evaluated the detection performance by calculating the area under the curve (AUC) of the receiver operating characteristic (ROC) curve. The AUC score, which ranges from 0 to 1, reflects the model's ability to distinguish anomalous pixels from the background at various threshold settings, with a higher AUC indicating better performance. To assess the performance, we compared our results with several existing methods: Reed-Xiaoli (RX) \cite{reed1990adaptive}, unsupervised ensemble-kernel principal component analysis (UE-KPCA) \cite{merrill2020unsupervised}, low-rank and sparse representation (LRASR) \cite{xu2015anomaly}, robust graph autoencoders (RGAE) \cite{9494034}, pixel-shuffle downsampling blind-spot reconstruction network (PDBSNet) \cite{wang2023pdbsnet}, and blind-block reconstruction network (BockNet) \cite{wang2023bocknet}.

Fig.~\ref{fig:roc_avirisI}–\ref{fig:roc_forest} present the ROC curves and AUC scores for the proposed method and competing approaches across five datasets: AVIRIS-I, AVIRIS-II, HYDICE Urban, Pavia University, and HYDICE Forest Radiance I. In each figure, the left panel shows the full ROC curves for all methods, with AUC scores computed by integrating up to false positive rate (FPR) $ = 10^{0}$. These results indicate that the proposed method consistently outperforms the competing HAD algorithms across all datasets when considering the full FPR range.

However, in many practical scenarios, detection performance under low FPRs is of greater importance. The right panels of Fig.~\ref{fig:roc_avirisI}–\ref{fig:roc_forest} display the partial ROC curves restricted to $\text{FPR} \leq 10^{-2}$. The corresponding AUC scores demonstrate that the proposed method achieves superior performance at low FPRs on four of the five datasets—AVIRIS-I, AVIRIS-II, HYDICE Urban, and Pavia University. On the HYDICE Forest Radiance I dataset, however, the proposed method's AUC under $\text{FPR} \leq 10^{-2}$ is slightly lower than those of LRASR and RX. These findings are further supported by Table~\ref{tab:auc_scores}, which reports comparative AUC scores integrated up to $\text{FPR} = 10^{-3}$, $10^{-2}$, and $10^{0}$. The results confirm that, except for the Forest dataset at low FPR thresholds, the proposed method consistently outperforms the competing approaches. Notably, for the Forest dataset, the proposed method achieves the highest AUC when considering the entire ROC curve.

\begin{figure}[tb!]
\begin{minipage}[b]{\linewidth}
  \centering
  \centerline{\includegraphics[width=0.95\textwidth]{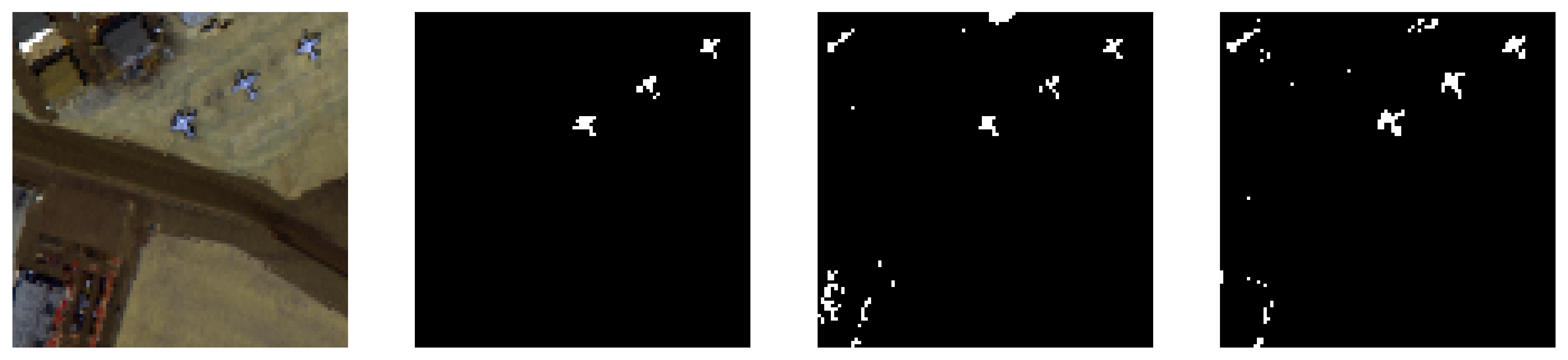}}
  \centerline{(a) AVIRIS-I}\medskip
\end{minipage}
\begin{minipage}[b]{\linewidth}
  \centering
  \centerline{\includegraphics[width=0.95\textwidth]{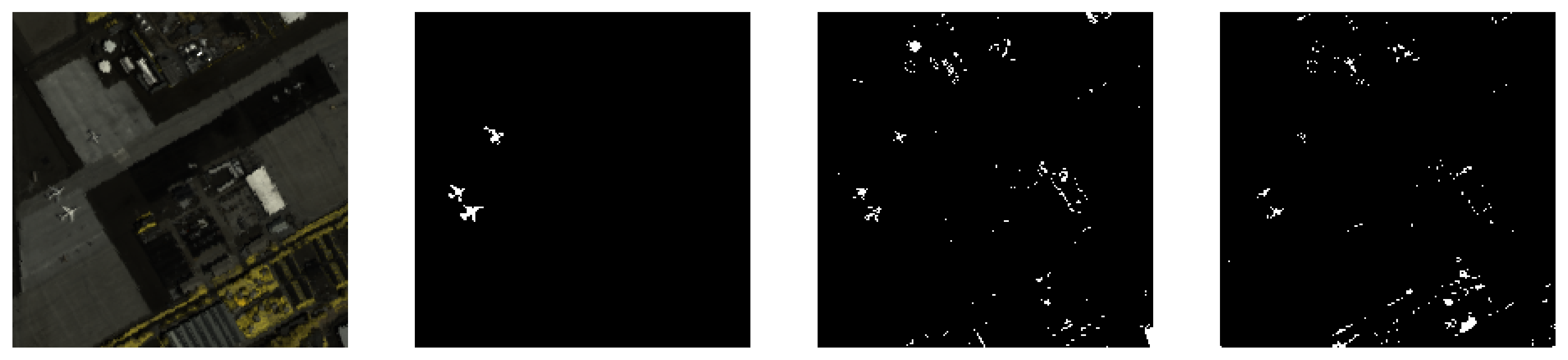}}
  \centerline{(b) AVIRIS-II}\medskip
\end{minipage}
\hfill
\begin{minipage}[b]{\linewidth}
  \centering
  \centerline{\includegraphics[width=0.95\textwidth]{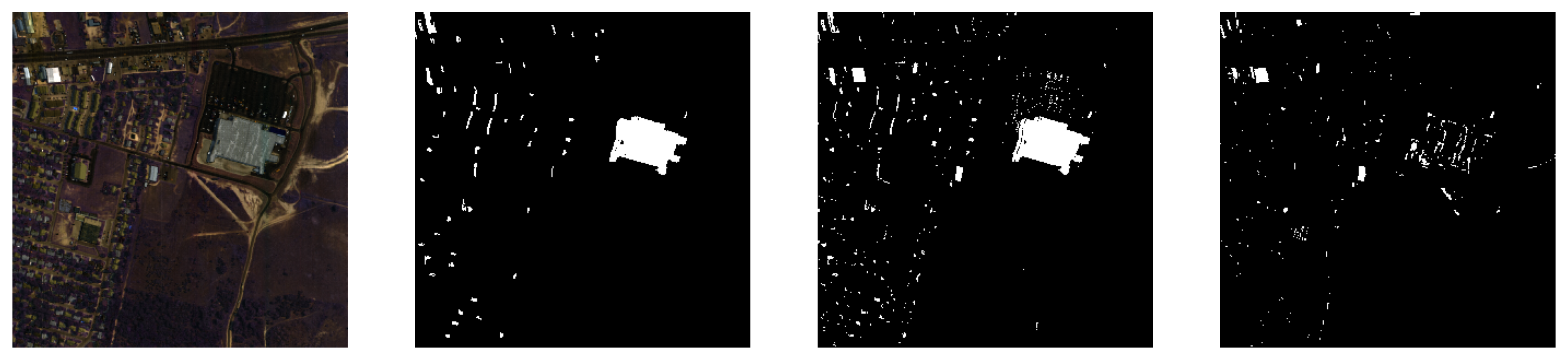}}
  \centerline{(c) Urban}\medskip
\end{minipage}
\begin{minipage}[b]{\linewidth}
  \centering
  \centerline{\includegraphics[width=0.95\textwidth]{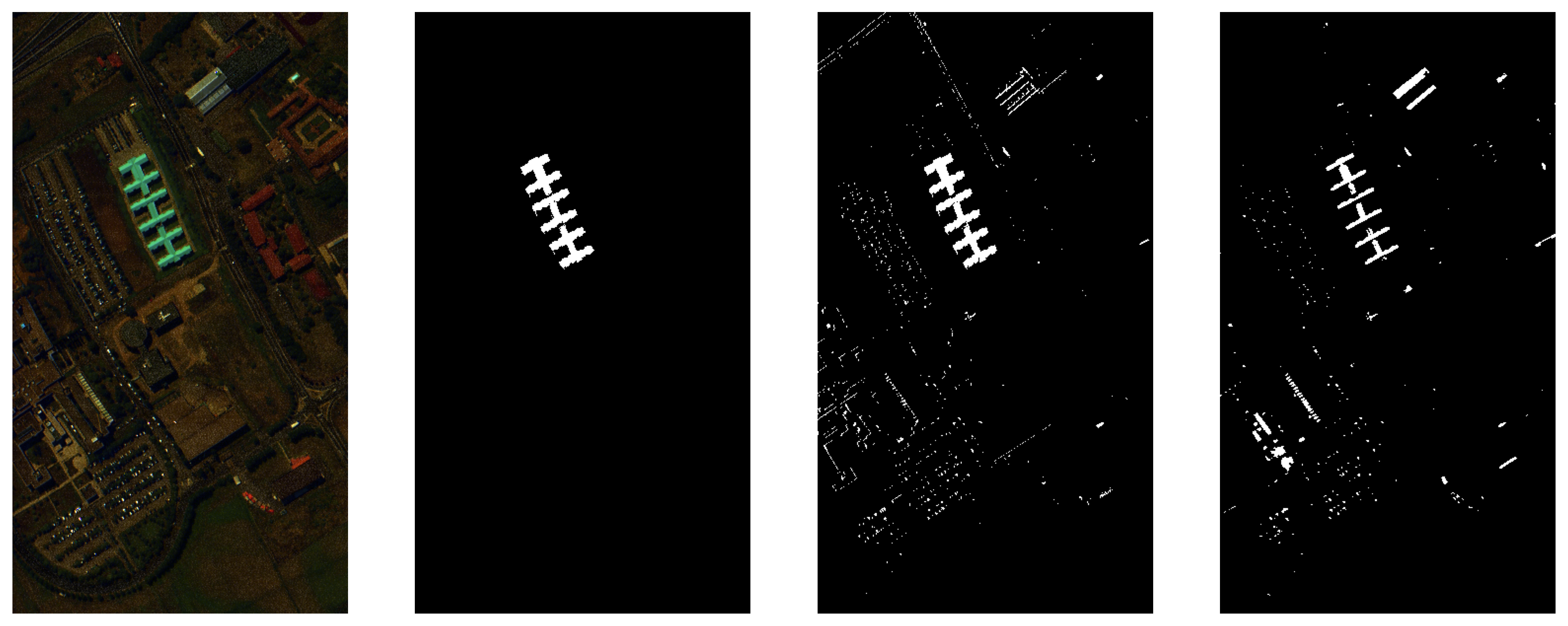}}
  \centerline{(d) Pavia}\medskip
\end{minipage}
\hfill
\begin{minipage}[b]{\linewidth}
  \centering
  \centerline{\includegraphics[width=0.95\textwidth]{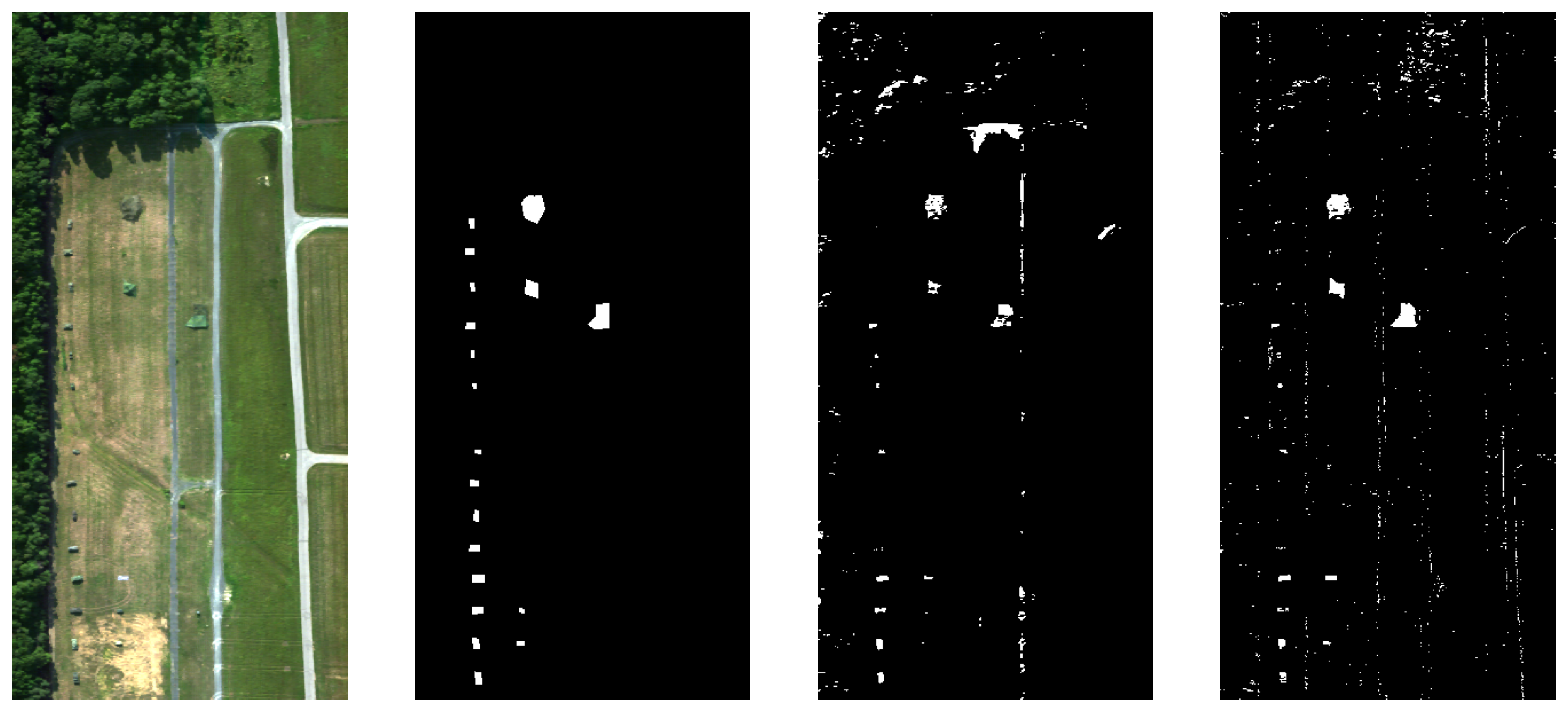}}
  \centerline{(e) Forest}\medskip
\end{minipage}
\caption{Each subfigure presents, from left to right: approximate RGB image of the dataset, corresponding ground-truth map, binary detection map generated by the proposed method with a specific threshold that yields $FPR=10^{-2}$, and binary detection map obtained by the best competing method with threshold corresponding to $FPR=10^{-2}$. The best competing methods for the datasets AVIRIS-I, AVIRIS-II, Urban, Pavia, and Forest are BockNet, PDBSNet, UE-KPCA, RGAE, and LRASR, respectively.}
\label{fig:bin_datasets}
\end{figure}

Fig.~\ref{fig:bin_datasets} displays the binary detection maps corresponding to thresholds set to achieve a FPR of $10^{-2}$ for each dataset. For each subfigure, the third column shows the detection result produced by the proposed method, while the fourth column presents the result of the best-performing competing method for that dataset. Visual inspection shows that the proposed method consistently yields more accurate anomaly detections compared to the competing methods, with the exception of the Forest Radiance I dataset. 
The relatively suboptimal performance on the Forest dataset can be attributed to the model assumption in Eq.~(\ref{eq:bg_model}), which presumes that the hyperspectral image is largely homogeneous except for sparse anomalies. While this assumption generally holds for the dataset, the presence of tree shadows combined with a small road segment introduces significant inhomogeneity. As a result, the proposed method misclassifies these regions as anomalies (see the third column of Fig.~\ref{fig:bin_datasets}(e)). 
In contrast, for the AVIRIS-I, AVIRIS-II, Urban, and Pavia datasets (Fig.~\ref{fig:bin_datasets}(a)–(d)), the proposed method demonstrates superior anomaly detection and reduced false alarms compared to the best competing methods. These qualitative results are consistent with the quantitative performance observed in the ROC curves (Figs.~\ref{fig:roc_avirisI}–\ref{fig:roc_forest}) and further reinforce the effectiveness of the proposed approach, particularly in scenarios that conform to the underlying model defined in Eq.~(\ref{eq:bg_model}). 

\section{Summary and Conclusion}
\label{sec:conc}
Unsupervised hyperspectral anomaly detection (HAD) has emerged as a powerful tool for applications ranging from environmental monitoring to defense, owing to its ability to identify spectral irregularities without prior knowledge of targets. Traditional model-based approaches, such as the RX algorithm and its variants, provide statistically principled frameworks, but they often struggle with the complexity of background distributions in real-world applications. Representation-based methods provide greater flexibility by leveraging reconstruction strategies, though they can be computationally intensive and parameter-dependent. More recently, deep learning-based approaches have shown strong empirical performance, particularly with autoencoders and self-supervised methods, yet they still face challenges in interpretability and theoretical grounding. Collectively, these developments highlight both the progress and the remaining challenges in building robust, efficient, and explainable HAD methods.

In this work, we propose a novel approach for hyperspectral anomaly detection by introducing a transport-based mathematical model for the background pixels of hyperspectral images. The HAD problem is first formulated on the basis of this model, and an unsupervised, noniterative solution is then proposed using a subspace modeling technique in the SCDT domain. Although problem formulation and solution are based on the assumption that background signals are observations of a template pattern under confounding deformations, knowledge of this template or deformations is not required. Extensive experimental results demonstrated the superior performance of our method over existing approaches, making it well suited for real-world applications. Future research will focus on adapting this approach to handle more complex scenarios, where the background of the HSI may not fully conform to the proposed model.




\bibliographystyle{IEEEtran}
\bibliography{refs}

\newpage

 




\vfill

\end{document}